\documentclass{article}
\usepackage{spconf,amsmath,graphicx}
\usepackage{times}
\usepackage{epsfig}
\usepackage{amsmath}
\usepackage{amssymb}
\usepackage{mathtools}
\usepackage{multirow}
\usepackage{xcolor}

\newcommand\blfootnote[1]{%
  \begingroup
  \renewcommand\thefootnote{}\footnote{#1}%
  \addtocounter{footnote}{-1}%
  \endgroup
}

\def\etal{et al. }

\def\baseline{S2VT}

\def\multibaseline{IMGD}
\def\methodsrcmsa{SRCMSA}

\usepackage{bbding}

\title{Video Question Generation via Semantic Rich Cross-Modal Self-Attention Networks Learning}
\name{Yu-Siang Wang*$\dag$, Hung-Ting Su*$\ddagger$, Chen-Hsi Chang$\ddagger$, Zhe-Yu Liu$\ddagger$, Winston H. Hsu$\ddagger$}
\address{$\dag$University of Toronto, $\ddagger$National Taiwan University }

\begin{document}
%

\maketitle
\blfootnote{*: Both authors contributed equally to this research.}
\begin{figure*}
  \includegraphics[width=\textwidth]{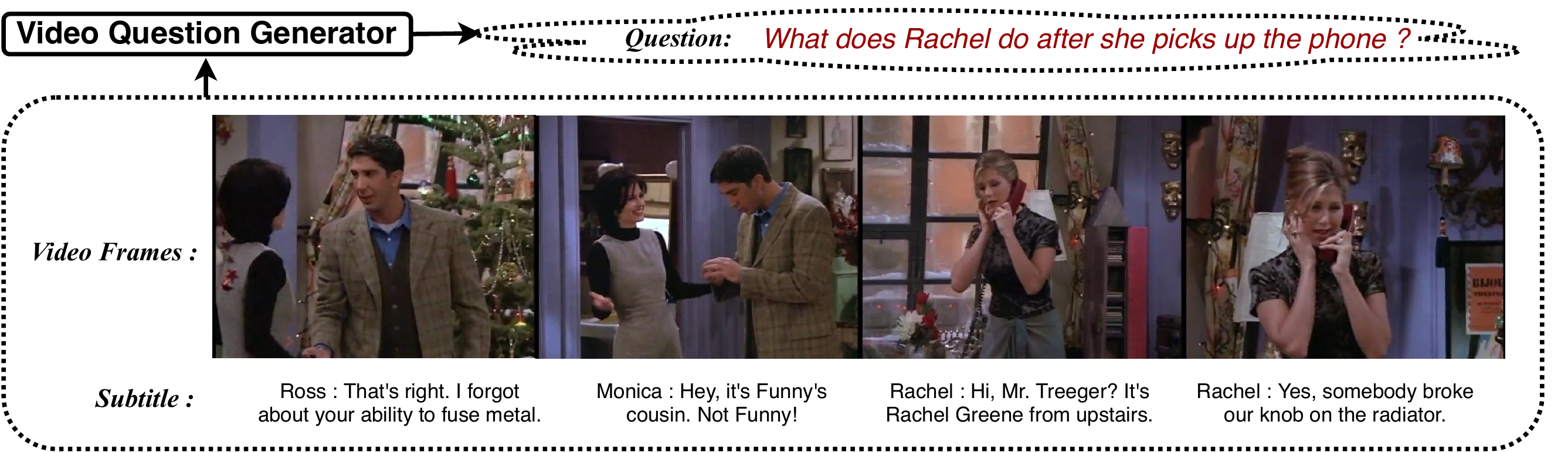}
  \caption{Video Question Generation (Video QG). We propose a novel and challenging task which automatically generates an answerable question (Top) according to a video clip (Bottom). Video QG model is required to represent the sparsity of the video features but also understand the interaction between objects and the semantic information of the dialogues.}
  
  \label{fig:task}
\end{figure*}
\begin{abstract}
%
%
We introduce a novel task, Video Question Generation (Video QG). A Video QG model automatically generates questions given a video clip and its corresponding dialogues. Video QG requires a range of skills -- sentence comprehension, temporal relation, the interplay between vision and language, and the ability to ask meaningful questions. To address this, we propose a novel semantic rich cross-modal self-attention (SRCMSA) network to aggregate the multi-modal and diverse features. To be more precise, we enhance the video frames semantic by integrating the object-level information, and we jointly consider the cross-modal attention for the video question generation task. Excitingly, our proposed model remarkably improves the baseline from 7.58 to 14.48 in the BLEU-4 score on the TVQA dataset. Most of all, we arguably pave a novel path toward understanding the challenging video input and we provide detailed analysis in terms of diversity, which ushers the avenues for future investigations.
\end{abstract}
\section{Introduction}
\label{sec: Intro}
Recent years have witnessed the rise of interest in vision-language tasks due to daily generated multimedia contents from the world. Among these tasks, video understanding tasks have drawn researchers' attention. However, all existing tasks mainly focus on learning to `answer' instead of learning to `ask', which is also equivalently crucial to society. For instance, online teaching videos on the MOOC or the BBC English Learning website are required to ask questions to evaluate the students' understanding of the video materials. In addition, we humans learn by asking questions of the parts we are not confident of. As for a robot, it can learn faster by asking questions about the learning experience which it is not able to understand. Shen et al.\cite{Shen19} improved the image captioning model by allowing the model to ask questions but their method is limited to static image inputs. Moreover, the ability to ask can also aid the development in popular vision-language fields, such as Visual Dialogue task\cite{visual_dialogue}, Video Question Answering\cite{tvqa}, and Embodied Question Answering\cite{embodiedqa} by decreasing the expensive questions annotation cost. 

To achieve these, we propose a novel and practical task, Video Question Generation, as illustrated in Figure \ref{fig:task}. A successful Video QG model should ask a meaningful question based on a video clip. Comparing to the text question generation task \cite{L2A} where the target questions are highly overlapped with the input passage or the image question generation task \cite{DBLP:conf/cvpr/LiDZCOWZ18} where the questions are regardless of the temporal information, Video QG is arguably more challenging. To tackle Video QG, it is intuitive to apply
LSTM based Seq2Seq models \cite{S2VT, DBLP:conf/naacl/VenugopalanXDRM15, yao2015capgenvid} to our task. However, there are several problems to be addressed. (1) The visual features are extracted by pre-trained convolutional neural networks (CNNs) and frozen during training. This limited the features to incorporate the relationships between multiple objects or between objects and the scene. (2) Multimedia information fusion mechanism for LSTM models usually compute multimodal attention with only one embedding space for a single modality. This restricted the fusion representation from learning the multiple semantic subspaces and the intersection between different subspaces of different modalities.

Therefore, we propose the following novel approaches to cope with the aforementioned problems:
(1) \textbf{Semantic-Rich Embedding (SRE)}. 
Instead of directly feeding the CNN extracted features or object layouts \cite{obj2text} to the model, we formalize the scene and objects relationship by incorporating the object-level semantic meaning into the visual information via fusing the CNN-extracted features and object-level representation. 
(2) \textbf{Cross-Modal Self-Attention (CMSA) encoder}. Our proposed CMSA encoder could solve the notorious long-term dependencies issues by applying the self-attention mechanism. In addition, we learn multiple semantic subspaces for each modality, which allows the model to capture complementary semantic attention in different subspaces.


We train our system on the TVQA \cite{tvqa} dataset, the only
available dataset that is human-labeled and is labeled by having people watch attentively on both visual and textual data of real and untrimmed videos, which is the best testbed for our Video QG model. We apply the Video QG in two scenarios, video frames and video frames with dialogues, as these two scenarios are both common in multimedia contents. 
Our proposed model significantly surpasses the competitive baseline models based on previous works \cite{imgd,S2VT, obj2text} on BLEU, BLEU-4, ROUGE, CIDEr, and METEOR scores in both scenarios. In addition, Video QG requires the model to generate a good question in terms of correctness, diversity, and answerability. Hence, we also carefully analyze the diversity and answerability of generated questions. The results not only demonstrate the effectiveness of our methods but provide a path for future researches and applications.
%


\section{Semantic Rich Cross-Modal Self-Attention Network}\label{section: selfmodel}

\begin{figure*}
  \includegraphics[width=\textwidth]{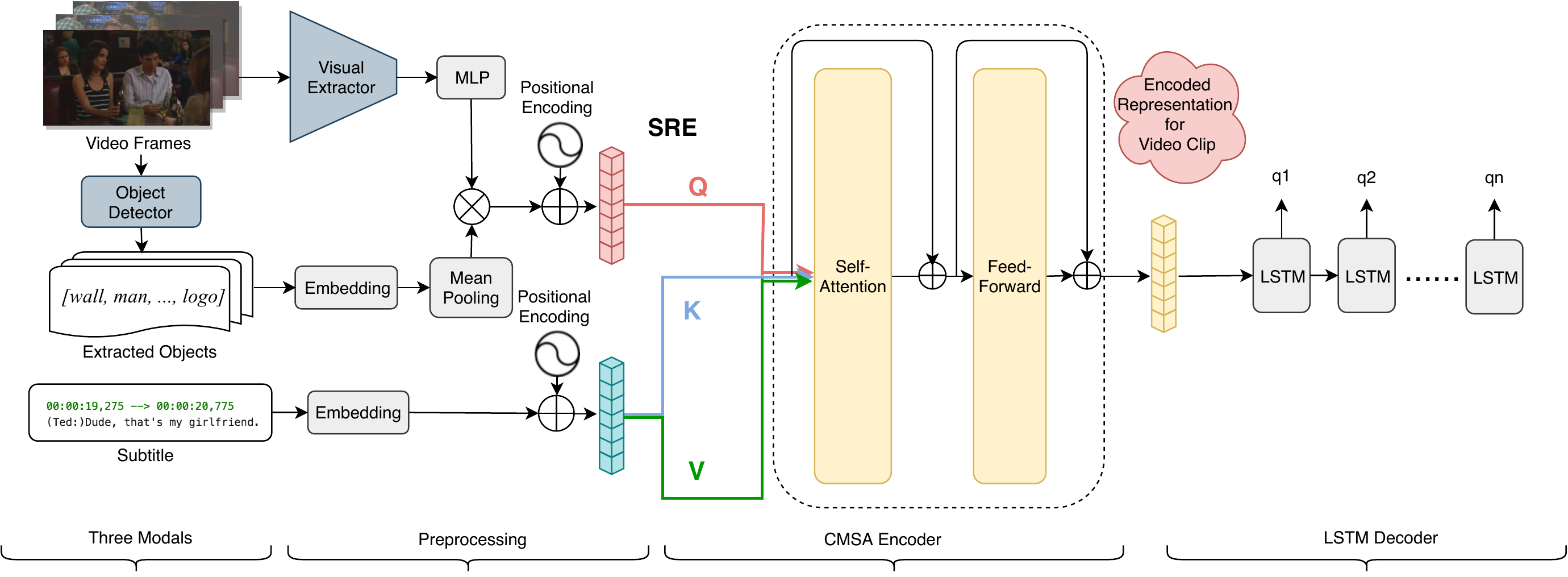}
  \caption{The framework of our proposed semantic rich cross-modal self-attention (SRCMSA) networks. The architecture consists of two transformers as the encoder and an LSTM as the decoder. Our proposed SRCMSA networks enable more rich interactions between videos (e.g., attentive semantic rich embedding) and the dialogues, which are complementary to each other (cf. section \ref{section: selfmodel}).} 
  \label{fig:architecture}
\end{figure*}

As shown in Figure \ref{fig:architecture}, our novel SRCMSA networks take multi-modal inputs. First, we represent the visual features with SRE to fuse the comprehensive object cues with CNN features. Then, we aggregate the multi-modal information from the video frames and dialogues by CMSA. Finally, an LSTM decoder is applied to generate the desired question.
\subsection{Features}
Our model inputs multi-modal features including (1) CNN extracted visual feature ${V}^{frame}  \{{V}^{F}_{1}, {V}^{F}_{2}, \cdots , {V}^{F}_{n_{frame}}\}$, $V^{F}_{i} \in{\mathbb{R}}^{2048}$, (2) Object features 
${V}^{O}  \{{V}^{O}_{1}, {V}^{O}_{2}, \dots , {V}^{O}_{n_{frame}}\}$, where $V^{O}_{i} \in{\mathbb{R}}^{300}$ , and (3) Dialogue features \newline ${V}^{sub}  \{{V}^{sub}_{1}, {V}^{sub}_{2}, \dots , {V}^{sub}_{n_{S}}\}$, ${V}^{sub}_{i} \in{\mathbb{R}}^{300}$, where ${n}_{S}$ is the word length of the dialogues in a video clip. The features can be obtained in various ways. In our experiments, we extract video frames from the video in 3 FPS, then get CNN features from the Pool5 layer of ResNet101 \cite{resnet} and detect objects features by Faster-RCNN \cite{DBLP:journals/pami/RenHG017}. The dialogue features are from the dialogues in the video. Both of the detected object in the frames and the word in dialogues are then embedded with 300 dimension vectors. Afterward, we mean pool the objects representation in each frame and get object embeddings.

\subsection{Semantic Rich Cross-Modal Self-Attention Encoder}\label{section: crossmodal}
We start introducing our novel CMSA encoder. We solve the long-range dependencies problem by utilizing the self-attention mechanism and aggregate the fusion subspaces by running multi-head attention of different modalities. Overall, we first compute our proposed semantic rich embedding (SRE) then incorporate the SRE with the dialogues with our novel cross-modal mechanism.

\subsection{Semantic Rich Embedding (SRE)}\label{section: sre}
SRE captures the essential cues in sparse video input by fusing visual ResNet features ${V}^{F}$ and comprehensive object occurrence features ${V}^{O}$ into joint representation. First, we obtain the video feature for each frame:
\begin{equation}
    V^{src}_{i} = W_{proj}(V^{F}_{i}) \odot V^{O}_{i}
\end{equation}
where the visual feature is projected to the embedding space by a matrix $W_{proj}$, then a dot product is applied to combine both visual features and object occurrence features.
 
\subsection{Cross-Modal Self-Attention (CMSA) Mechanism}\label{cmsa} 
\textbf{Single Head Attention}.
Transformer\cite{transformer} has achieved huge success in many areas. However, most of the existing works limit the transformer for single modality input. In our proposed CMSA, we aim to integrate the rich information from multi modalities. We first begin by introducing the single head attention used in our CMSA:
\begin{equation}
Attention(Q, K, V) = Attention({W}^{Q}Q, {W}^{K}K, {W}^{V}V)
\end{equation}
$Attention(Q, K, V)=softmax(\frac{QK}{\sqrt{d}})V $, where Q, K, V, d are served as the query, key, value, and the dimension of the input. $W^{Q},W^{K}, W^{V}$ are the projected matrices. \newline  We set the SRE features as our query, and the dialogues as the key and value because the dialogues features are able to provide richer semantic information than the visual features.
\newline
\textbf{Multi-Head Attention}.
A single attention matrix computation is called a ``head", and we adopt the \textit{mutli-head attention} in our model. The multi-head attention consists of \textit{H} parallel heads which allow the model to jointly learn from different representation subspaces. The multi-head attention output is computed as below:
\begin{align}\label{eq:multihead}
& {head}_{j} = Attention({W}_{j}^{Q}Q, {W}_{j}^{K}K, {W}_{j}^{V}V) \\
 & MultiHead(Q, K, V) = {W}^{O}Concat({head}_{1},...,{head}_{H}) 
\end{align}

Corresponding to our experiment, a head computation can be viewed as a fusion between the projected subspaces from each input modality.  Overall, our cross-modal attentive value ${V}^{cmsa}$ is then calculated by:
\begin{align}
    {V}^{cmsa} & = {MultiHead}({{V}^{sub}}^{Q}, {{V}^{sub}}^{K}, {{V}^{Sre}}^{V})  \\ & = {MultiHead}({W}^{Q}{V}^{sub}, {W}^{K}{V}^{sub}, {W}^{V}{V}^{sre})
\end{align}
Every project matrix , ${W}^{Q}_{j}$,  ${W}^{K}_{j}$, or ${W}^{V}_{j}$, is a learnable weight and indicates a different semantic subspace projection which allows the model to aggregate the fusion from different attention subspaces.

\begin{table*}
    \centering
    \begin{tabular}{
    |llc|
    c|c|c|c|c|
    }
    \hline
    
    & & w/ Sub  
    & BLEU
    & BLEU-4
    & ROUGE
    & CIDEr
    & METEOR
    \\ \hline

    \baseline{} \cite{S2VT}
    & &
    & 57.80 & 7.58 & 36.25 & 6.39 & 14.83 
    \\
    OBJ2TEXT \cite{obj2text}
    & &
    & 61.78 & 10.44 & 38.49 & 6.42 & 15.33
    \\
    \multibaseline{} \cite{imgd}
    & & \Checkmark
    & 61.08 & 9.59 & 37.78 & 7.29 & 15.21
    \\
    \hline
    SRCMSA (Ours) & (SA)
    &
    & 63.02 & 12.20 & 40.21 & 15.25 & 17.36 
    \\ 
    SRCMSA (Ours) & (SA + SRE)
    &
    & 66.26 & 13.74 & 41.48 & 22.93 & 18.66
    \\
    SRCMSA (Ours) & (SA + SRE + CMSA)
    & \Checkmark
    & \textbf{68.83} & \textbf{14.48} & \textbf{42.56} & \textbf{23.44} & \textbf{19.04}
    \\
    \hline

    \end{tabular}
    \caption{Video QG Results on TVQA \cite{tvqa} dataset. Sub: subtitles. SA: Self-Attention. We compare our proposed method with various competitive baselines (First Block). \cite{S2VT, obj2text, imgd}.
    SRE: Fusing the video frame feature with detected object level embedding mentioned in section \ref{section: crossmodal}. SRCMSA: our proposed model discussed in section \ref{section: selfmodel}. Our proposed method significantly outperforms the baselines. Ablation study in the second block demonstrates the effectiveness of both of our novel components. Noted that
    SRCMSA (w/o SRE, CMSA) is a single source model which utilize video frames as our input modality.
    }
    \label{table:results}
\end{table*}

%
%
\section{Experiments}\label{section:exp}
We evaluate our model on the TVQA dataset proposed by \cite{tvqa} Lei \etal. Most of the earlier datasets are either generated with templates \cite{tgqa} or annotated with single modality \cite{DBLP:conf/cvpr/TapaswiZSTUF16}. TVQA is based on 6 popular
TV shows and consists of 152,545 QA pairs from 21,793 clips. The input data involves videos and dialogues, and the crowd-workers are encouraged to annotate the questions requiring both modalities. Two scenarios are applied in terms of different multimedia model inputs: (1) Video only and (2) Video with dialogues. We quantitatively examine the correctness and the diversity of generated questions.


\begin{table*}\label{table:div1}
    \centering
    \begin{tabular}{
    |lll|
    c|c|c|
    c|c|c|
    c|c|c|
    c|c|c|
    }
    \hline
     & &
     & \multicolumn{3}{|c|}{Unigram (\%)}
     & \multicolumn{3}{|c|}{Bigram (\%)}
     & \multicolumn{3}{|c|}{Noun (\%)}
     & \multicolumn{3}{|c|}{Verb (\%)}
    \\
    &  &
    & 0.1\%
    & 1\%
    & 10\%
    & 0.1\%
    & 1\%
    & 10\%
    & 0.1\%
    & 1\%
    & 10\%
    & 0.1\%
    & 1\%
    & 10\%
    \\ \hline
    
    & \baseline{} \cite{S2VT}
    &
    & 72.2 & 99.9 & 100 
    & 87.8 & 100 & 100
    & 61.3 & 100 & 100
    & 39.0 & 99.9 & 100
    \\
    

    & \multibaseline{}  \cite{imgd}
    &
    & 34.5 & 99.9 & 100 
    & 68.0 & 100 & 100
    & 57.6 & 100 & 100
    & \textbf{31.9} & 99.9 &100
    \\

    \hline
    &  SRCMSA (Ours) 
    &
    & \textbf{33.8} & \textbf{87.9} & \textbf{98.0} 
    & \textbf{48.6} & \textbf{81.1} & \textbf{97.9}
    & \textbf{44.2} & \textbf{88.9} & 100
    & 49.4 & \textbf{99.5} & 100
    \\
    \hline
    
    \hline
    & GT (Oracle)
    &
    & 25.4 & 66.3 & 86.4
    & 16.4 & 38.7 & 64.7
    & 31.6 & 49.2 & 79.5
    & 16.9 & 57.0 & 84.2
    \\
    \hline
    \end{tabular}
    \caption{Frequent Word Coverage for Unigram, Bigram, Noun and Verb. The lower the more diverse. First block: Baselines. Second block: Our SRCMSA Model. Last block: Human labeled ground truth questions as the oracle case.
    Our model significantly surpasses the strong baseline model in terms of diversity.}
    \label{table:divbig}
\end{table*}

\subsection{Correctness}\label{section:correctness_result}
Table \ref{table:results} shows results of correctness evaluated by BLEU, BLEU-4, ROUGE, CIDEr, and METEOR metrics. Our method obtains an impressive gain across all metrics, indicating that our approach better utilizes the sparse and huge input video and dialogue features. Our ablation experiments (Second block) indicates the effectiveness of all proposed modules. Our CMSA encoder leverages the dialogue features in different domain while SRE represents the sparse video frame via using the object-level and frame-level signals.

\subsection{Diversity}\label{section:diversity_result}
Table \ref{table:divbig} demonstrates diversity by employing the ratio of frequent words coverage. Our model outperforms both baselines in a significant margin. Additionally, the ground truth questions annotated by humans are much more diverse, indicating that the Video QG is a challenging task. Also, besides the correctness metrics which are widely applied, we argue that diversity measurement deserves attention.

\section{Analysis and Future WOrk}\label{section:future_direction}

We identify 3 common errors in the results, including \textbf{unanswerable}, \textbf{redundant} and \textbf{general} questions. We desire to point out some future directions to overcome these errors in the further research.

There are two main types of unanswerable questions: (1) \textit{Action Error} where the question refers to nonexistent action. (2) \textit{Entity Error} where the involved entity does not exist. Using action recognition models or entity recognition models should be helpful to mitigate the issues. Also, the researchers can include specific answers as the queries to the Video QG model in the training stage to generate answerable questions.

\textbf{Redundant} questions are answerable but contain unnecessary conditions. Such as ``wearing a shirt" in the question ``Who is the person wearing a shirt" when there is only one person in the video. In future works, we are able to decide which questions are redundant with the aid of referring expression models \cite{refexpression}. 

\textbf{General} questions are answerable questions which can be applied to any video. Such as ``What's the person doing". General questions are ``correct" but suboptimal because they're simple and discourage reasoning during QA process. General questions also harm the diversity results in Table \ref{table:divbig}. We suggest the researchers to generate more specific questions by following the strategies in \textbf{Unanswerable} error.

\section{Conclusion}
In this paper, we offer a new perspective and solution to video understanding by proposing a novel task, Video QG, which automatically generates the questions given a video clip. We propose the SRCMSA model to represent and understand the video properly by attending on multimodal sequential features.
Our model significantly surpasses the competitive baseline networks, \baseline, \multibaseline, in terms of quantity and quality. In addition to evaluating with the commonly used metrics, we propose the frequent word coverage to measure the diversity of the generated questions. We dig into the quality beyond the scores by analyzing errors. 
We hope our work can lead to more thoughts on the creative uses and extensions of Video QG. In the future, we plan to augment Video QA dataset with our improved Video QG model, such as the insufficient Video QA dataset of the languages other than English.

\section*{Acknowledgement}
This work was supported in part by the Ministry of Science and Technology, Taiwan, under Grant MOST 109-2634-F-002-032. We benefit from the NVIDIA grants and the DGX-1 AI Supercomputer and are also grateful to the National Center for High-performance Computing.

\clearpage
\bibliographystyle{IEEEbib}
\bibliography{strings,refs}

\begin{thebibliography}{10}

\bibitem{Shen19}
Kevin Shen, Amlan Kar, and Sanja Fidler,
\newblock ``Lifelong learning for image captioning by asking natural language
  questions,''
\newblock in {\em ICCV}, 2019.

\bibitem{visual_dialogue}
Dat~Tien Nguyen, Shikhar Sharma, Hannes Schulz, and Layla~El Asri,
\newblock ``From film to video: Multi-turn question answering with multi-modal
  context,''
\newblock {\em CoRR}, vol. abs/1812.07023, 2018.

\bibitem{tvqa}
Jie Lei, Licheng Yu, Mohit Bansal, and Tamara Berg,
\newblock ``Tvqa: Localized, compositional video question answering,''
\newblock in {\em Proceedings of the 2018 Conference on Empirical Methods in
  Natural Language Processing}. 2018, pp. 1369--1379, Association for
  Computational Linguistics.

\bibitem{embodiedqa}
Abhishek Das, Samyak Datta, Georgia Gkioxari, Stefan Lee, Devi Parikh, and
  Dhruv Batra,
\newblock ``{E}mbodied {Q}uestion {A}nswering,''
\newblock in {\em Proceedings of the IEEE Conference on Computer Vision and
  Pattern Recognition (CVPR)}, 2018.

\bibitem{L2A}
Xinya Du, Junru Shao, and Claire Cardie,
\newblock ``Learning to ask: Neural question generation for reading
  comprehension,''
\newblock in {\em Proceedings of the 55th Annual Meeting of the Association for
  Computational Linguistics (Volume 1: Long Papers)}. 2017, pp. 1342--1352,
  Association for Computational Linguistics.

\bibitem{DBLP:conf/cvpr/LiDZCOWZ18}
Yikang Li, Nan Duan, Bolei Zhou, Xiao Chu, Wanli Ouyang, Xiaogang Wang, and
  Ming Zhou,
\newblock ``Visual question generation as dual task of visual question
  answering,''
\newblock in {\em 2018 {IEEE} Conference on Computer Vision and Pattern
  Recognition, {CVPR} 2018, Salt Lake City, UT, USA, June 18-22, 2018}, 2018,
  pp. 6116--6124.

\bibitem{S2VT}
Subhashini Venugopalan, Marcus Rohrbach, Jeffrey Donahue, Raymond Mooney,
  Trevor Darrell, and Kate Saenko,
\newblock ``Sequence to sequence -- video to text,''
\newblock in {\em Proceedings of the 2015 IEEE International Conference on
  Computer Vision (ICCV)}, Washington, DC, USA, 2015, ICCV '15, pp. 4534--4542,
  IEEE Computer Society.

\bibitem{DBLP:conf/naacl/VenugopalanXDRM15}
Subhashini Venugopalan, Huijuan Xu, Jeff Donahue, Marcus Rohrbach, Raymond~J.
  Mooney, and Kate Saenko,
\newblock ``Translating videos to natural language using deep recurrent neural
  networks,''
\newblock in {\em {HLT-NAACL}}. 2015, pp. 1494--1504, The Association for
  Computational Linguistics.

\bibitem{yao2015capgenvid}
Li~Yao, Atousa Torabi, Kyunghyun Cho, Nicolas Ballas, Christopher Pal, Hugo
  Larochelle, and Aaron Courville,
\newblock ``Describing videos by exploiting temporal structure,''
\newblock in {\em Computer Vision (ICCV), 2015 IEEE International Conference
  on}. IEEE, 2015.

\bibitem{obj2text}
Xuwang Yin and Vicente Ordonez,
\newblock ``Obj2text: Generating visually descriptive language from object
  layouts,''
\newblock in {\em Proceedings of the 2017 Conference on Empirical Methods in
  Natural Language Processing}, Copenhagen, Denmark, Sept. 2017, pp. 177--187,
  Association for Computational Linguistics.

\bibitem{imgd}
Iacer Calixto and Qun Liu,
\newblock ``Incorporating global visual features into attention-based neural
  machine translation,''
\newblock in {\em Proceedings of the 2017 Conference on Empirical Methods in
  Natural Language Processing, {EMNLP} 2017, Copenhagen, Denmark, September
  9-11, 2017}, 2017, pp. 992--1003.

\bibitem{resnet}
Kaiming He, Xiangyu Zhang, Shaoqing Ren, and Jian Sun,
\newblock ``Deep residual learning for image recognition,''
\newblock in {\em 2016 {IEEE} Conference on Computer Vision and Pattern
  Recognition, {CVPR} 2016, Las Vegas, NV, USA, June 27-30, 2016}, 2016, pp.
  770--778.

\bibitem{DBLP:journals/pami/RenHG017}
Shaoqing Ren, Kaiming He, Ross~B. Girshick, and Jian Sun,
\newblock ``Faster {R-CNN:} towards real-time object detection with region
  proposal networks,''
\newblock {\em {IEEE} Trans. Pattern Anal. Mach. Intell.}, vol. 39, no. 6, pp.
  1137--1149, 2017.

\bibitem{transformer}
Ashish Vaswani, Noam Shazeer, Niki Parmar, Jakob Uszkoreit, Llion Jones,
  Aidan~N. Gomez, Lukasz Kaiser, and Illia Polosukhin,
\newblock ``Attention is all you need,''
\newblock in {\em {NIPS}}, 2017, pp. 6000--6010.

\bibitem{tgqa}
Yunseok Jang, Yale Song, Youngjae Yu, Youngjin Kim, and Gunhee Kim,
\newblock ``{TGIF-QA:} toward spatio-temporal reasoning in visual question
  answering,''
\newblock in {\em 2017 {IEEE} Conference on Computer Vision and Pattern
  Recognition, {CVPR} 2017, Honolulu, HI, USA, July 21-26, 2017}, 2017, pp.
  1359--1367.

\bibitem{DBLP:conf/cvpr/TapaswiZSTUF16}
Makarand Tapaswi, Yukun Zhu, Rainer Stiefelhagen, Antonio Torralba, Raquel
  Urtasun, and Sanja Fidler,
\newblock ``Movieqa: Understanding stories in movies through
  question-answering,''
\newblock in {\em {CVPR}}. 2016, pp. 4631--4640, {IEEE} Computer Society.

\bibitem{refexpression}
Junhua Mao, Jonathan Huang, Alexander Toshev, Oana Camburu, Alan~L. Yuille, and
  Kevin Murphy,
\newblock ``Generation and comprehension of unambiguous object descriptions,''
\newblock in {\em 2016 {IEEE} Conference on Computer Vision and Pattern
  Recognition, {CVPR} 2016, Las Vegas, NV, USA, June 27-30, 2016}, 2016, pp.
  11--20.

\end{thebibliography}

\end{document}